\documentclass[journal]{IEEEtran}
\ifCLASSINFOpdf
\else
\fi

\usepackage{times}
\usepackage{epsfig}
\usepackage{graphicx}
\usepackage{float}
\usepackage{amsmath}
\usepackage{amssymb}
\usepackage{subfigure}
\usepackage{overpic}
\usepackage{multirow}
\usepackage{booktabs}
\usepackage{enumitem}
\usepackage{setspace}
\usepackage{threeparttable}

\usepackage{algpseudocode}
\usepackage{algorithmicx,algorithm}

\begin{document}
%
\title{CFNet: LiDAR-Camera Registration Using Calibration Flow Network}
%
%
%

\author{Xudong~Lv,~Boya~Wang,~Dong~Ye,~Ziwen~Dou~and~Shuo~Wang
\thanks{Xudong Lv, Boya Wang, Ziwen Dou, Dong Ye, and Shuo Wang are with School of Instrumentation Science and Engineering, Harbin Institute of Technology, Harbin 150001, China email:(15B901019@hit.edu.cn; 19B901034@stu.hit.edu.cn; yedong@hit.edu.cn; 20B901041@stu.hit.edu.cn; 15B901018@hit.edu.cn).}}

\markboth{Journal of \LaTeX\ Class Files,~Vol.~14, No.~8, August~2015}%
{Shell \MakeLowercase{\textit{et al.}}: Bare Demo of IEEEtran.cls for IEEE Journals}
%



\maketitle

\begin{abstract}
As an essential procedure of data fusion, LiDAR-camera calibration is critical for autonomous vehicles and robot navigation. Most calibration methods rely on hand-crafted features and require significant amounts of extracted features or specific calibration targets. With the development of deep learning (DL) techniques, some attempts take advantage of convolutional neural networks (CNNs) to regress the 6 degrees of freedom (DOF) extrinsic parameters. Nevertheless, the performance of these DL-based methods is reported to be worse than the non-DL methods. This paper proposed an online LiDAR-camera extrinsic calibration algorithm that combines the DL and the geometry methods. We define a two-channel image named calibration flow to illustrate the deviation from the initial projection to the ground truth. EPnP algorithm within the RANdom SAmple Consensus (RANSAC) scheme is applied to estimate the extrinsic parameters with 2D-3D correspondences constructed by the calibration flow. Experiments on KITTI datasets demonstrate that our proposed method is superior to the state-of-the-art methods. Furthermore, we propose a semantic initialization algorithm with the introduction of instance centroids (ICs). The code will be publicly available at https://github.com/LvXudong-HIT/CFNet.

\end{abstract}


%
\IEEEpeerreviewmaketitle


\section{Introduction}
%
%
%
%
\IEEEPARstart{S}{cene} perception is an essential aspect of autonomous driving and robot navigation. Robust perception of the objects in the surrounding environment relies on various onboard sensors installed on the mobile platform. Light Detection and Ranging (LiDAR) sensors can obtain the spatial measurement of the scene with a wide range and high accuracy. But LiDAR sensors have a low resolution, especially on the horizontal orientation, and lack color and texture information. Camera sensors can provide high-resolution RGB images but are sensitive to illumination changes and have no distance information. The combination of LiDAR and camera sensors can make up for each others' shortcomings. The effective fusion of these two sensors is critical to the 3D object detection \cite{chen2017multi, ku2018joint, qi2018frustum, xu2018pointfusion}, and semantic mapping tasks \cite{jeong2018multimodal, yue2020collaborative, li2020building}. As an essential process to LiDAR-camera fusion, extrinsic calibration should be applied to estimate the rigid body transformation between LiDAR and the camera accurately.

\begin{figure}[thpb]
    \centering
    \includegraphics[width=8.5cm]{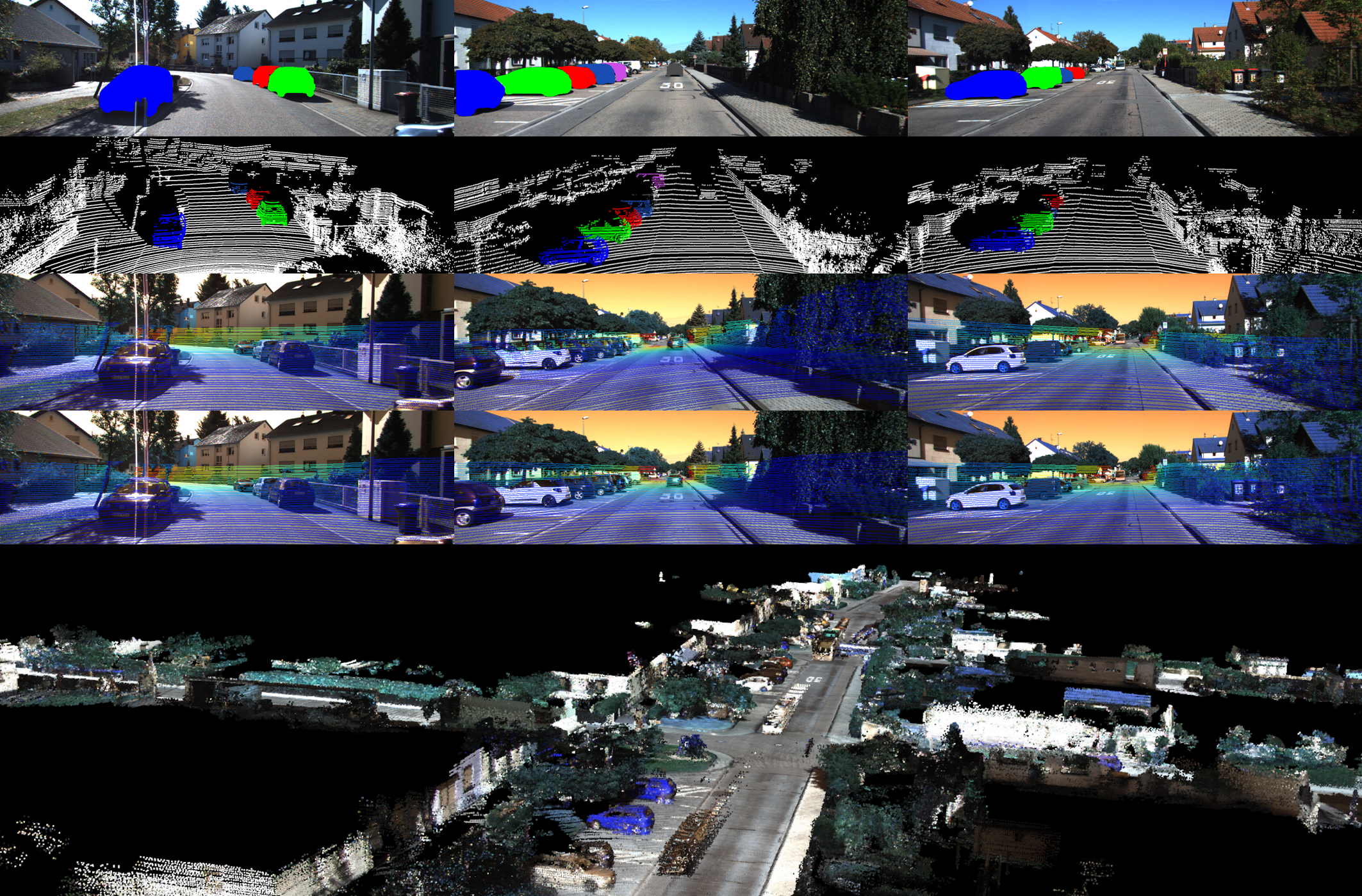}
    \caption{Examples of our proposed CFNet. (First Row) The 2D instance centroid (2D-IC) extracted from images. (Second Row) The 3D instance centroid (3D-IC) extracted from LiDAR point clouds. (Third Row) The initial calibration parameters provided by semantic initialization (registration of 2D-IC and 3D-IC). (Fourth Row) The final calibration parameters predicted by CFNet. (Bottom Row) Three-dimensional map generated by fusing the sensors using calibration values provided by our approach.}
    \label{CFNet show}
\end{figure}

Early LiDAR-camera calibration methods relied on specific calibration targets, such as a checkerboard or self-made objects. The matching relationship between 2D feature points (camera images) and 3D feature points (LiDAR point clouds) is obtained by manual labeling or automatic labeling, which is then used to calculate the extrinsic parameters. During the operation of an autonomous driving vehicle, the change of the operating environment (such as vibration, etc.) will introduce uncontrollable deviation to the extrinsic parameters. A series of targetless and online self-calibration algorithms were proposed to solve this problem, effectively improving adaptability. These online calibration methods mainly utilize the intensity of image and point clouds and the edge correlation to calculate the extrinsic calibration parameters. Such techniques rely on accurate initial extrinsic parameters or additional motion information (hand-eye calibration initialization). Besides, the features used to construct the penalty function of calibration parameters are usually set manually, highly dependent on the scene, and cannot work in the absence of features.

Recently, some approaches have attempted to apply deep learning to predict the relative transformation between LiDAR and the camera. The calibration errors of these methods are still significant compared with the traditional methods. Thus, these learning-based calibration methods can not meet the calibration requirements. Besides, in practical application, when sensor parameters change, many training data are needed to fine-tune the network model, which dramatically limits the algorithm's generalization.

In this paper, we propose an automatic online LiDAR-camera self-calibration method CFNet. Without the need for specific calibration scenes, calibration targets, and the initial calibration parameters, CFNet is fully automatic. We define a calibration flow to represent the deviation between the positions of initial projected points and the ground truth. The predicted calibration flow is utilized to rectify the initial projected points for accurate 2D-3D correspondences construction. The EPnP algorithm \cite{lepetit2009epnp} is adopted to calculate the final extrinsic parameters within RANSAC scheme \cite{fischler1981random}. To make CFNet fully automatic, we also propose a semantic initialization algorithm. More specifically, we showcase the contribution of CFNet as follows:
\begin{enumerate}
    \item To the best of our Knowledge, CFNet is the first LiDAR-camera extrinsic calibration approach that combines the deep learning and geometry method. Compared with the direct prediction of calibration parameters, CFNet has a stronger generalization.
    \item We define a calibration flow to predict the deviation between the positions of initial projected points and the ground truth. The calibration flow can shift the initial projection to the right positions construct accurate 2D-3D correspondences.
    \item We introduce the 2D/3D instance centroid (IC) to estimate the initial extrinsic parameters. With this semantic initialization procedure, CFNet can execute LiDAR-camera calibration automatically.
    \item The source code is publicly available.
\end{enumerate}

\section{Related Work}
According to whether requiring specific calibration targets, the LiDAR-camera extrinsic calibration algorithms can be categorized into target-based and targetless. Target-based methods usually require calibration targets in an empty scene and estimate the extrinsic parameters by extracting the features from the targets. Targetless approaches extract hand-crafted feature points from the environment to construct a penalty function for the calibration parameters. Given an accurate initial value, the optimal extrinsic parameters are obtained by optimizing the penalty function. Deep learning-based approaches are proposed in recent years. These methods use the convolutional neural networks (CNNs) to replace the traditional feature extraction, feature matching, and initialization for extrinsic parameters prediction.

\subsection{Target-based Approach}
The checkerboards are mainly used to calibrate cameras and are also suitable for the LiDAR-camera calibration \cite{zhang2004extrinsic, kassir2010reliable, an2020geometric, zhou2012extrinsic, kim2020extrinsic}. The checkboard's pose in camera coordinate was calculated by triangulating the checkerboard's corner points extracted from the image. In the LiDAR point clouds, the checkerboard was identified by the segmentation method and establish the correlation between the plane points of LiDAR and the camera. With multiple checkerboards \cite{geiger2012automatic}, an error function containing correlation constraints of multiple plane points can be established. To ensure the algorithm's robustness, the checkerboards need to be evenly distributed in the sensor's field of view (Angle and position). Verma \emph{et al.} \cite{verma2019automatic} extracted the central coordinates and the corresponding plane normal vector of the checkerboard as matching features. A genetic algorithm (GA) \cite{mitchell1998introduction} was applied to acquire a globally optimal calibration result.

A systematic range of reflection bias is often observed in the LiDAR scanning results on the checkerboards, which will lead to measurement errors and affect the final calibration results. Park \emph{et al.} \cite{park2014calibration} used a monochromatic board as the calibration target instead of the checkerboard to deal with this issue. Dhall \emph{et al.} \cite{dhall2017lidar} added the ArUco markers on the plate as the calibration target. Benefit from the ArUco markers, the LiDAR-camera calibration task was transformed into a 3D-3D ICP problem. Guindel \emph{et al.} \cite{guindel2017automatic} designed a custom-made calibration target that hollows out four circles on a rectangular board. The stereo camera and LiDAR detected the circles and fit the central coordinates of the four circles respectively for the estimation of rigid body transformation. Beltr{\'a}n \emph{et al.} \cite{beltran2021automatic} pasted the ArUco markers on the calibration plate proposed in \cite{guindel2017automatic}, which makes the new calibration target has the ability to calibrate LiDAR and monocular cameras. The LiDAR's resolution is much lower than that of the camera, so the calibration methods based on the detection of the plane edge are restricted by the accuracy of LiDAR edge extraction.

Besides planar calibration targets, spherical targets are also appropriate for LiDAR-camera calibration \cite{kummerle2018automatic, kummerle2020unified, toth2020automatic}. Compared with planar targets, the advantage of the spherical target is that the camera can automatically detect the outline without depending on the camera's angle of view and placement. Besides, sampling points of spherical objects can be detected conveniently on the LiDAR point cloud \cite{toth2020automatic}.

\subsection{Targetless Approach}
During the long-life operation of robots, the extrinsic parameters inevitably deviate to some extent. To ensure a long-term stable operation, the robot needs to have the ability to detect the deviation and rectify the bias of the extrinsic parameters. Online calibration algorithms without specific calibration targets can effectively solve this problem. 

Depth discontinuous points in the LiDAR point clouds are corresponding to the edge points in the image. When project the LiDAR point clouds to the image plane given correct extrinsic parameters, the 3D points with depth discontinuity will be more likely to be projected onto the edge of the image \cite{castorena2016autocalibration}. In this hypothesis, Levinson \emph{et al.} \cite{levinson2013automatic} proposed an online self-calibration method that contains an online deviation detection module and rectification module. \cite{pandey2012automatic} introduced the information theory into the calibration task. The rigid body transformation parameters are estimated by analyzing the statistical correlation between LiDAR and camera measurements. Mutual information (MI) was selected as the measurement metric of statistical correlation. Besides the 3D LiDAR point clouds and RGB images, the reflection intensity of LiDAR is also used to construct the MI objective function. Taylor \emph{et al.} \cite{taylor2013automatic} leveraged a new measurement parameter, Gradient Orientation Measure (GOM), to describe the gradient-related of LiDAT point clouds and images.

The above methods do not need calibration targets and are entirely data-driven calibration algorithms. Nevertheless, an appropriate initial calibration value is indispensable. A significant deviation between the initial value and the ground truth will bring about a wrong correlation. Therefore, these methods can only be used for fine-tuning calibration parameters. \cite{ishikawa2018lidar} transformed the LiDAR-camera calibration into an extended problem of solving a hand-eye calibration problem. Hand-eye calibration can provide initial extrinsic parameters, but it depends heavily on visual odometry and LiDAR odometry accuracy \cite{taylor2016motion}. \cite{shi2019extrinsic, zhen2019joint, park2020spatiotemporal} combined the LiDAR-camera calibration with sensor fusion localization and mapping to establish a joint optimization function.
The odometry and extrinsic parameters are optimized simultaneously for stable mapping. 

\subsection{Deep Learning Approach}
It is inspiring to note that deep learning technology has made a breakthrough in many fields such as classification, object detection, semantic segmentation, and object tracking. Some attempts apply deep neural networks to the multi-sensors calibration tasks. At present, the research in this field is still in the preliminary stage. To our knowledge,  RegNet \cite{schneider2017regnet} was the first deep learning method that transformed the feature extraction, feature matching, and global regression into real-time CNNs to deduce the 6-DOF of extrinsic parameters between LiDAR and camera. RegNet ignores the geometry nature of $SE(3)$ using the quaternion distance as training loss. CalibNet \cite{iyer2018calibnet} proposed a 3D spatial transformer layer (3D STL) to deal with this problem. The output of the 3D STL is used to re-project the LiDAR point clouds to formulate a geometric loss. An end-to-end training is performed by maximizing the geometric and photometric consistency between the input image and the point cloud. The above deep learning calibration methods ignore the tolerance within the error bounds. RGGNet \cite{yuan2020rggnet} utilized the Riemannian geometry and deep generative model to build a tolerance-aware loss function. 

Semantic information is introduced for obtaining an ideal initial extrinsic parameter. SOIC \cite{wang2020soic} exploited semantic information to calibrate and transforms the initialization problem into the Perspective-n-Points (PnP) problem of the semantic centroid. Due to the 3D semantic centroid of the point cloud and the 2D semantic centroid of the image cannot match accurately, a matching constraint cost function based on the semantic elements of the image and the LiDAR point cloud is also proposed. The optimal calibration parameter is obtained by minimizing the cost function. Zhu \emph{et al.} \cite{zhu2020online} proposed an online calibration system that automatically calculates the optimal rigid-body motion transformation between two sensors by maximizing the mutual information of their perceived data without adjusting the environment Settings. By formulating the calibration as an optimization problem with semantic features, the temporally synchronized LiDAR and camera are registered in real-time.

\section{Method}
In this section, we first describe the definition of the calibration flow. We then present our proposed calibration method CFNet, including the network architecture, loss functions, semantic initialization, and the training details.

\begin{figure*}[thpb]
    \centering
    \includegraphics[width=17cm]{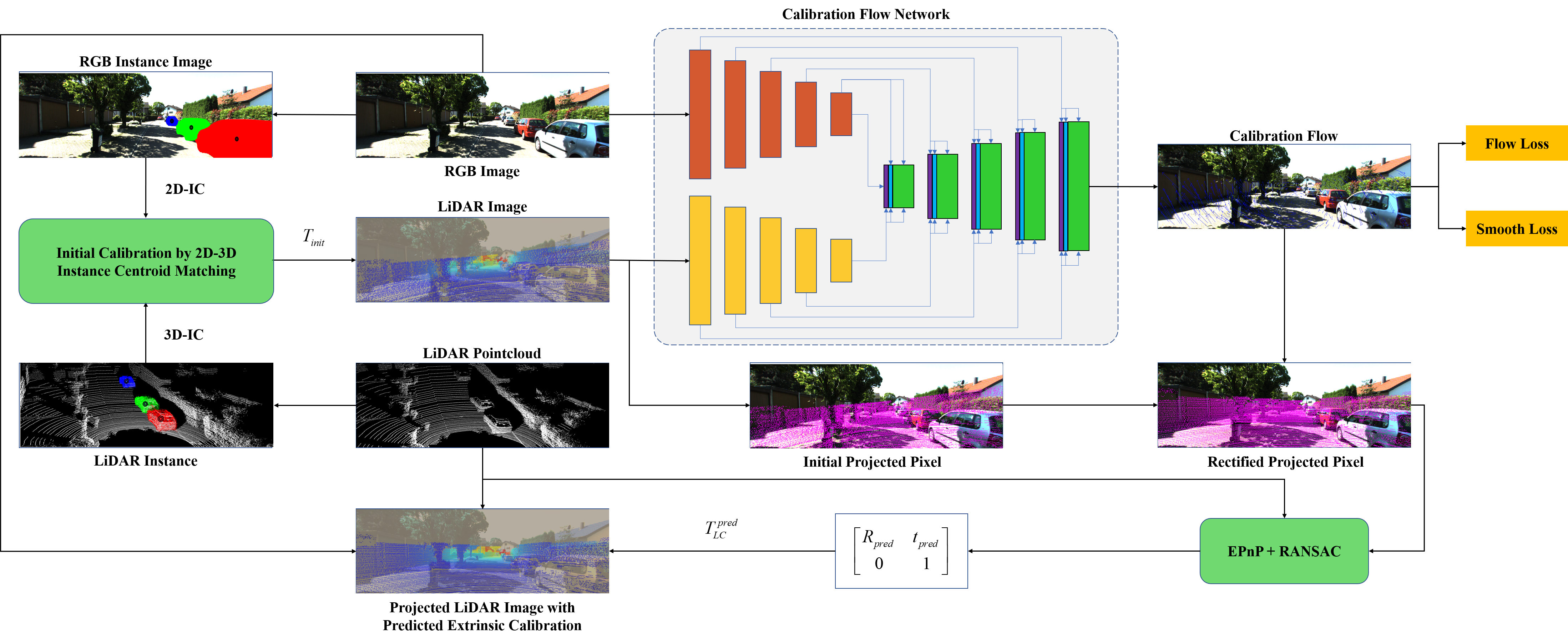}
    \caption{The workflow of our proposed CFNet, an automatic online extrinsic calibration method that estimates the transformation parameters between 3D LiDAR and 2D camera.}
    \label{CFNet workflow}
\end{figure*}

\subsection{Calibration Flow}
Given initial LiDAR-Camera extrinsic parameters ${T}_{init}$ and the camera intrinsic matrix $K$, each LiDAR 3D point cloud $P_{i}^{L}={{[\begin{matrix}
   X_{i}^{L} & Y_{i}^{L} & Z_{i}^{L} \end{matrix}]}^{T}}\in {{\mathbb{R}}^{3}}$ is projected onto the image plane with corresponding 2D pixel coordinate $p_{i}^{init}={{\left[ \begin{matrix}
   u_{i}^{init} & v_{i}^{init} \end{matrix} \right]}^{T}}\in {{\mathbb{R}}^{2}}$
. This projection process is expressed as follows:
\begin{equation}
    \begin{aligned}
      Z_{i}^{init}\cdot \hat{p}_{i}^{init} & =Z_{i}^{init}\left[ \begin{matrix}
       u_{i}^{init}  \\
       v_{i}^{init}  \\
       1  \\
    \end{matrix} \right] =K\left[ {{R}_{init}}|{{t}_{init}} \right]\hat{P}_{i}^{L} \\ 
     & =K\left[ {{R}_{init}}|{{t}_{init}} \right]\left[ \begin{matrix}
       X_{i}^{L}  \\
       Y_{i}^{L}  \\
       Z_{i}^{L}  \\
       1  \\
    \end{matrix} \right] \\ 
    \end{aligned}
\end{equation}

\begin{equation}
    {{T}_{init}}=\left[ \begin{matrix}
    {{R}_{init}} & {{t}_{init}}  \\
    0 & 1  \\
    \end{matrix} \right]
\end{equation}

where $\hat{P}_{i}^{L}$ is the homogeneous coordinates of $P_{i}^{L}$, $\hat{p}_{i}^{init}$ is the homogeneous coordinates of $p_{i}^{init}$. The projected LiDAR-image ${{D}_{init}}$ is acquired by z-buffer approach, each pixel $(u_{i}^{init},v_{i}^{init})$ in the image reserves the depth value $Z_{i}^{init}$.
\begin{equation}
    {{D}_{init}}\left( u_{i}^{init},v_{i}^{init} \right)=Z_{i}^{init},i\in {{VI}_{init}}
\end{equation}

where ${VI}_{init}$ is the valid index of the initial 2D-3D correspondences, which fulfils $0<u_{i}^{init}<W,0<v_{i}^{init}<H$, $W$ and $H$ are width and height of the RGB image. If the pixel in ${{D}_{init}}$ does not have any matched LiDAR point cloud, we set this pixel value to 0. Similarly, we obtain the ground truth projected coordinate by projecting each 3D LiDAR point cloud with extrinsic ground truth ${T}_{gt}$ and camera intrinsic $K$.
\begin{equation}
    \begin{aligned}
    Z_{i}^{gt}\cdot \hat{p}_{i}^{gt} & =Z_{i}^{gt}\left[ \begin{matrix}
       u_{i}^{gt}  \\
       v_{i}^{gt}  \\
       1  \\
    \end{matrix} \right]=K\left[ {{R}_{gt}}|{{t}_{gt}} \right]\hat{P}_{i}^{L} \\ 
     & =K\left[ {{R}_{gt}}|{{t}_{gt}} \right]\left[ \begin{matrix}
       X_{i}^{L}  \\
       Y_{i}^{L}  \\
       Z_{i}^{L}  \\
       1  \\
    \end{matrix} \right] \\ 
    \end{aligned}
\end{equation}

\begin{equation}
    {{T}_{gt}}=\left[ \begin{matrix}
    {{R}_{gt}} & {{t}_{gt}}  \\
    0 & 1  \\
    \end{matrix} \right]
\end{equation}

where $\hat{p}_{i}^{gt}$ is the homogeneous coordinates of $p_{i}^{gt}={{\left[ \begin{matrix}
u_{i}^{gt} & v_{i}^{gt} \end{matrix} \right]}^{T}}\in {{\mathbb{R}}^{2}}$. Each pixel $(u_{i}^{gt},v_{i}^{gt})$ in the ground truth LiDAR-image ${D}_{gt}$ reserves the depth value $Z_{i}^{gt}$.
\begin{equation}
    {{D}_{gt}}\left( u_{i}^{gt},v_{i}^{gt} \right)=Z_{i}^{gt},i\in {{VI}_{gt}}
\end{equation}

where ${VI}_{gt}$ is the valid index of the ground truth 2D-3D correspondences, which fulfills $0<u_{i}^{gt}<W,0<v_{i}^{gt}<H$, $W$ and $H$ are width and height of the RGB image. If the pixel in ${{D}_{gt}}$ does not have any matched LiDAR point cloud, we set this pixel value to zero. 

We define the deviation between the initial projected 2D coordinate $p_{i}^{init}$ and the ground truth $p_{i}^{gt}$ as the Calibration Flow. The Calibration Flow is similar to the optical flow, which includes two channels and represents the deviation of $p_{i}^{init}$ and $p_{i}^{gt}$ on the direction of $x$ and $y$ respectively. The ground truth calibration flow $F_{gt}^{calib}$ is calculated as follows, and the invalid pixels are set to zero.
\begin{equation}\label{cf_gt}
    \begin{aligned}
      & F_{gt}^{calib}\left( u_{i}^{init},v_{i}^{init} \right)=\left[ \begin{matrix}
       u_{i}^{gt}-u_{i}^{init} & v_{i}^{gt}-v_{i}^{gt}  \\
    \end{matrix} \right], \\ 
     & i\in \left( V{{I}_{init}}\cdot V{{I}_{gt}} \right) \\ 
    \end{aligned}
\end{equation}

\subsection{Network Architecture}
\begin{figure*}[thpb]
    \centering
    \includegraphics[width=16cm]{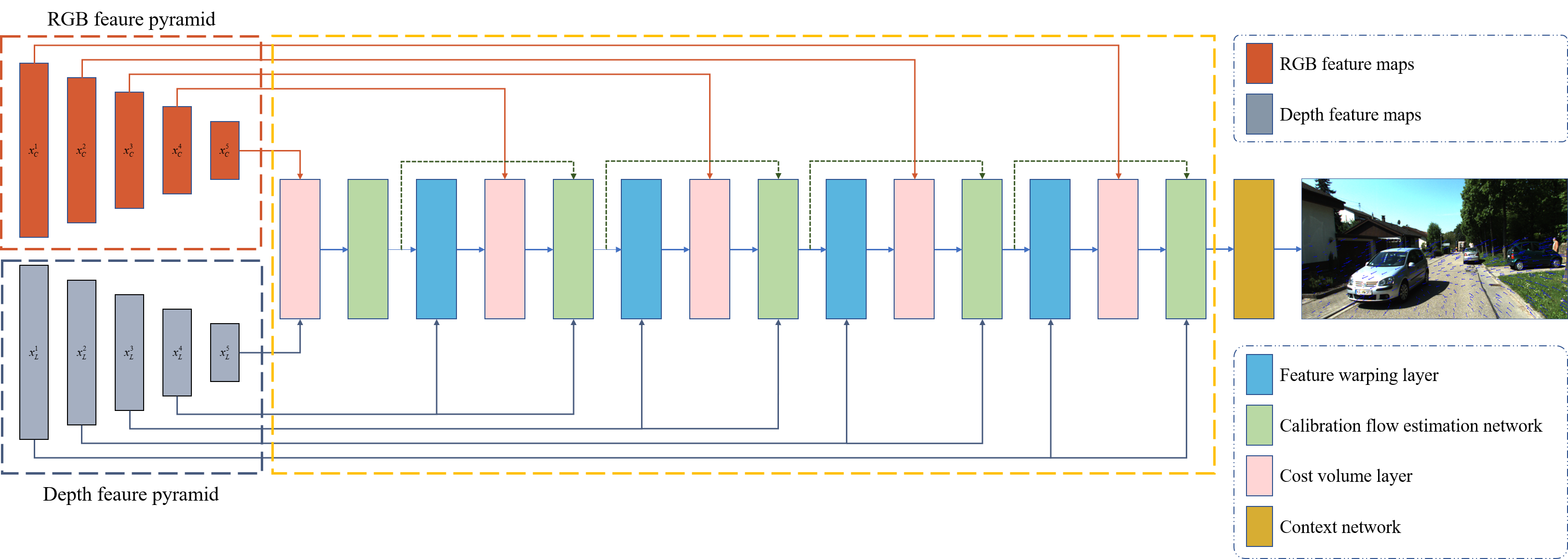}
    \caption{The Architecture of our proposed calibration network CFNet.}
    \label{CFNet Architecture}
\end{figure*}

We propose CFNet to predict the calibration flow, which is utilized to correct the initial projected 2D coordinate in the image plane. After coordinate shift using calibration flow, the accurate correspondences between 3D LiDAR point cloud and 2D projected point are constructed. As shown in Figure \ref{CFNet Architecture}, the architecture of CFNet is based on the PWC-Net \cite{sun2018pwc}, which is specifically designed for predicting the optical flow between adjacent frames. Unlike the original PWC-Net, we use a modified ResNet-18 architecture \cite{he2016deep} with the activation function leaky RELU in the Encoder Net.  The Encoder Net has two branches: the RGB branch and the Depth brach. Since the inputs of the two branches are heterogeneous images (RGB image and LiDAR point cloud), the weights are not shared between the two branch networks. Besides, since the LiDAR-image has only one channel, the number of input channels in the first convolution layer is 1, and the number of filters in all convolution layers is half of that in the RGB branch. The CFNet predicts an image $F_{pred}^{calib}$ with two channels, which represents the deviation of pixels between the initial projection $p_{i}^{init}$ and the ground truth $p_{i}^{gt}$.

\subsection{Loss Function}
We use two types of loss terms during training: the calibration flow photometric loss ${L}_{cf}$ and the calibration flow smooth loss ${L}_{s}$. 

(1) Calibration Flow Photometric Loss: 
After obtaining the predicted calibration flow, we inspect the sparse pixel-wise error between the predicted calibration flow $F_{pred}^{calib}$ and the ground truth calibration flow $F_{gt}^{calib}$ on valid pixels. We use the $L1$ norm for this photometric loss, and the error term is defined as,
\begin{equation}
    {{L}_{cf}}=\frac{\sum\limits_{u,v}{{{\left\| F_{gt}^{calib}(u,v)-F_{pred}^{calib}(u,v) \right\|}_{1}}\cdot M(u,v)}}{\sum\limits_{u,v}{M(u,v)}}
\end{equation}

\begin{equation}
    M(u_{i}^{init},v_{i}^{init})=\left\{ \begin{array}{l}
    1 \quad i\in \left( {{VI}_{init}}\cdot V{{VI}_{gt}} \right)  \\
    0 \quad otherwise
    \end{array} \right.
\end{equation}

where $M$ is the mask of valid pixels in $F_{gt}^{calib}$.

(2) Calibration Flow Smoothness Loss:
The smoothness loss ${{L}_{s}}$ of the calibration flow is different from the optical flow. For dense optical flow, the correspondence flow maps are encouraged to be locally smooth, making the values of adjacent pixels close. Our proposed calibration flow is sparse, and most of the pixels are invalid. Thus, the function of the smoothness loss ${{L}_{s}}$ is to enforce the displacement of pixels without a ground truth to be similar to the ones in the neighboring pixels.
\begin{equation}
    {{L}_{s}}=\frac{\sum\limits_{u,v}{{{D}_{s}}(u,v)\cdot (1-M(u,v))}}{\sum\limits_{u,v}{(1-M(u,v))}}
\end{equation}

\begin{equation}
    \begin{aligned}
    {{D}_{s}}(u,v) & =\rho (F_{pred}^{calib}(u,v)-F_{pred}^{calib}(u+1,v)) \\
                   & +\rho (F_{pred}^{calib}(u,v)-F_{pred}^{calib}(u,v+1))
    \end{aligned}
\end{equation}

where $\rho $ is the generalized Charbonnier function $\rho (x)=({{x}^{2}}+{{\varepsilon }^{2}}),\varepsilon ={{10}^{-9}},\alpha =0.25$, as in \cite{jason2016back}.

\subsection{Calibration Inference}
We make shift correction on the initial 2D projection point $p_{i}^{init}$ using calibration flow. The shift correction process is shown as follows,
\begin{equation}
    \begin{aligned}
        p_{i}^{rect} & =p_{i}^{init}+F_{pred}^{calib}(u_{i}^{init},v_{i}^{init}) \\
        & = \left[ \begin{matrix} u_{i}^{init} & v_{i}^{init} \end{matrix} \right]+F_{pred}^{calib}(u_{i}^{init},v_{i}^{init}),i\in V{{I}_{rect}}
    \end{aligned}
\end{equation}

\begin{equation}
    P_{i}^{L}=[\begin{matrix}
    X_{i}^{L} & Y_{i}^{L} & Z_{i}^{L}  \\
    \end{matrix}],i\in V{{I}_{rect}}
\end{equation}

where $p_{i}^{rect}=\left[ \begin{matrix} u_{i}^{rect} & v_{i}^{rect} \end{matrix} \right]\in {{\mathbb{R}}^{2}}$ is the rectified 2D coordinate of the valid projected LiDAR point cloud $P_{i}^{L}$, $V{{I}_{rect}}$ is the valid index of the 2D-3D correspondences which fulfil $0<u_{j}^{rect}<W,0<v_{j}^{rect}<H,0<u_{j}^{init}<W,0<v_{j}^{init}<H$. After acquiring a number of valid 2D-3D correspondences, we transfer the LiDAR-Camera extrinsic calibration task to a PnP problem. We use the EPnP algorithm within RANSAC scheme to solve this, with a maximum of 10 iterations, 5 times repeats, and an inlier threshold value of 1 pixel. 

Similar to LCCNet, we employ a multi-range iterative refinement method to improve the calibration accuracy further. We train five models with different initial error ranges $\Delta T$, $\left[ -x,x \right],x=\left\{ 1.5m,1.0m,0.5m,0.2m,0.1m \right\}$ for translation and $\left[ -y,y \right],y=\left\{ 20{}^\circ ,10{}^\circ ,5{}^\circ ,2{}^\circ ,1{}^\circ  \right\}$ for rotation. The calibration iterative refinement process is shown in Algorithm 1. The inputs are camera frame $I$, LiDAR point clouds $\left\{ P_{i}^{L} \right\}$, camera intrinsic $K$ as well as initial calibration parameters ${T}_{init}$. After processing the sensors data, we project the LiDAR point clouds $\left\{ P_{i}^{L} \right\}$ to image plane to generate the sparse depth image ${D}_{init}$ and the projected 2D points $\{p_{i}^{init}\}$. Due to the input size of the Network is $\text{320}\times \text{960}$, we need to crop the original RGB image and the depth image simultaneously. To ensure the cropped depth image contains as many points as possible, we calculate the centroid of $\{p_{i}^{init}\}$ to get the location of the crop window. Then, we use the output of ${{N}_{1}}(\pm 1.5m,\pm 20{}^\circ )$ to rectify the coordinate of each initial projected point. The rectified 2D projected points and the corresponding valid LiDAR point clouds construct new 2D-3D correspondences. By applying the EPnP algorithm within the RANSAC scheme, we calculate the extrinsic parameters ${T}_{pred}$ and set it to ${T}_{1}$. We regard the transformation ${T}_{1}$ as new initial extrinsic parameters ${T}_{init}$ to re-project the LiDAR point clouds and generate depth image, which is the input of ${{N}_{2}}(\pm 1.0m,\pm 10{}^\circ )$. The number of accurate 2D-3D correspondences in the current iteration is much more than the last iteration in theory. The above process is repeated five times, iteratively improving the extrinsic calibration accuracy. The final calibration result is represented by ${{\widehat{T}}_{LC}}={{T}_{5}}$.

\begin{algorithm}[t]
\setstretch{1.35}
\caption{Extrinsic Calibration Iterative Refinement} 
\hspace*{0.02in} {\bf Input:}
$I$, $\{P_{i}^{L}\}$, $K$, ${T}_{init}$, ${N}_{valid}$ \\
\hspace*{0.02in} {\bf Output:} 
${{\widehat{T}}_{LC}}$
\begin{algorithmic}[]
\For{$k=1$ to 5} 
    \State ${{D}_{init}},\{p_{i}^{init}\}={{f}_{proj}}(\{P_{i}^{L}\},K,{{T}_{init}})$
    \State Crop $I$ and ${D}_{init}$ to $320\times 960$ adaptively
    \State $F_{pred}^{calib}={{N}_{i}}({{I}^{crop}},D_{init}^{crop})$
    \State $P_{i}^{rect}=P_{i}^{init}+F_{pred}^{calib}(u_{i}^{init},v_{i}^{init}),i\in {{VI}_{rect}}$
    \If{${{VI}_{rect}}>{N}_{valid}$} 
        \State Construct matches $\left\{ p_{i}^{rect},P_{i}^{L} \right\},i\in {{VI}_{rect}}$
        \State Calculate ${{T}_{pred}}$ by EPnP within RANSAC scheme
        \State ${{T}_{init}}={{T}_{pred}},{{T}_{i}}={{T}_{pred}}$
    \Else
        \State \textbf{break}
    \EndIf
\EndFor
\State ${{\widehat{T}}_{LC}}={{T}_{5}}$
\end{algorithmic}
\end{algorithm}

The method we proposed above only uses one frame of RGB image LiDAR point cloud to predict the extrinsic calibration parameters. In practical applications, inaccurate calibration flow will result in false calibration parameters. If we analyze the results over a sequence using the median value as the reference, the abnormal values that are significantly different from the reference can be eliminated.

\subsection{Semantic Initializtion}
Currently, most online calibration methods require an initialization process to provide a coarse value for optimizing the objective function. Manual measurement and hand-eye calibration are the two most commonly used initialization methods. However, the manual way is labor-intensive, and the hand-eye calibration method relies on accurate motion information. Inspired by SOIC \cite{wang2020soic}, we also try to use the semantic information to calculate the initial calibration parameters. 

\begin{figure}[thpb]
    \centering
        \subfigure[]{\includegraphics[width=4.2cm]{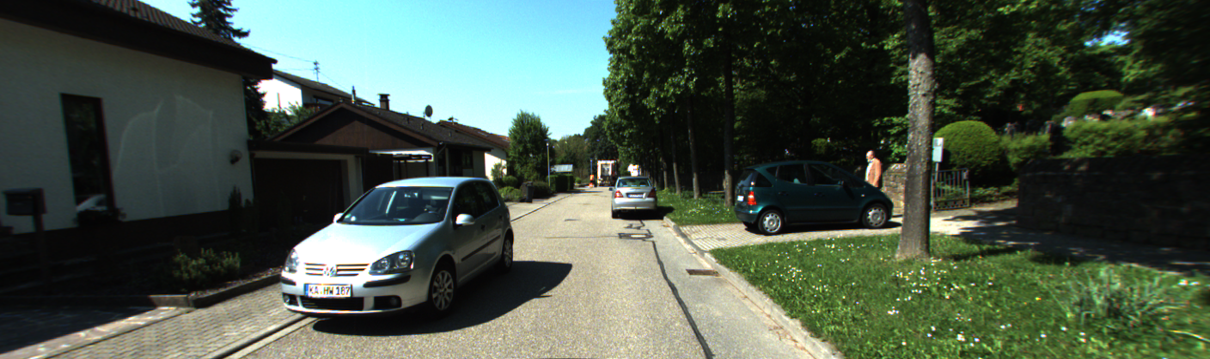}} 
        \subfigure[]{\includegraphics[width=4.2cm]{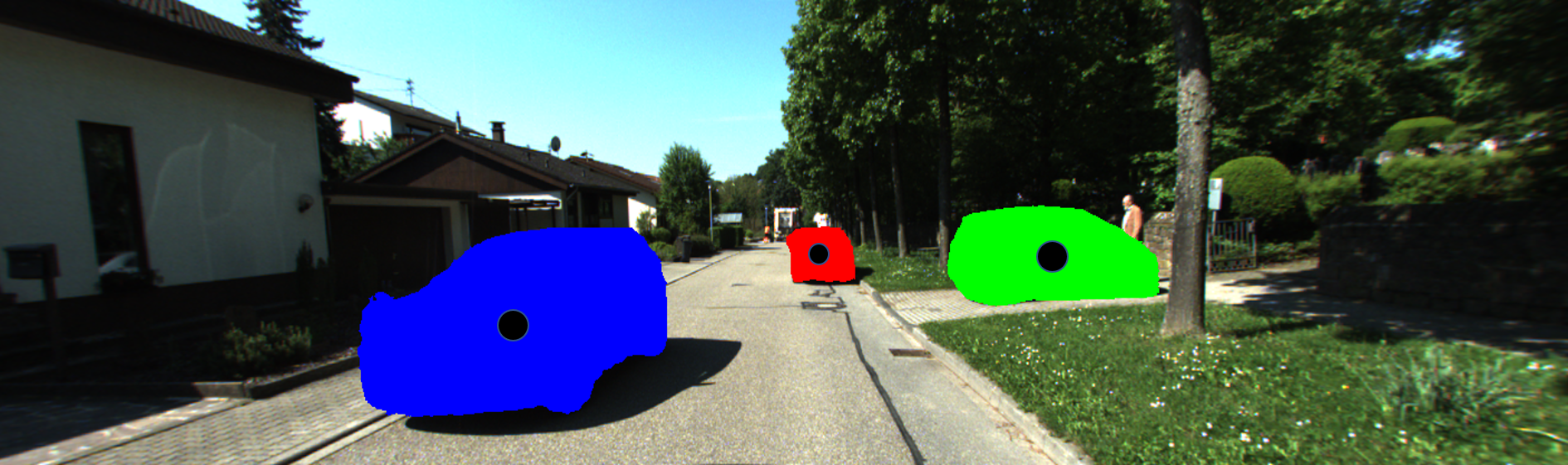}} \\
        \subfigure[]{\includegraphics[width=4.2cm]{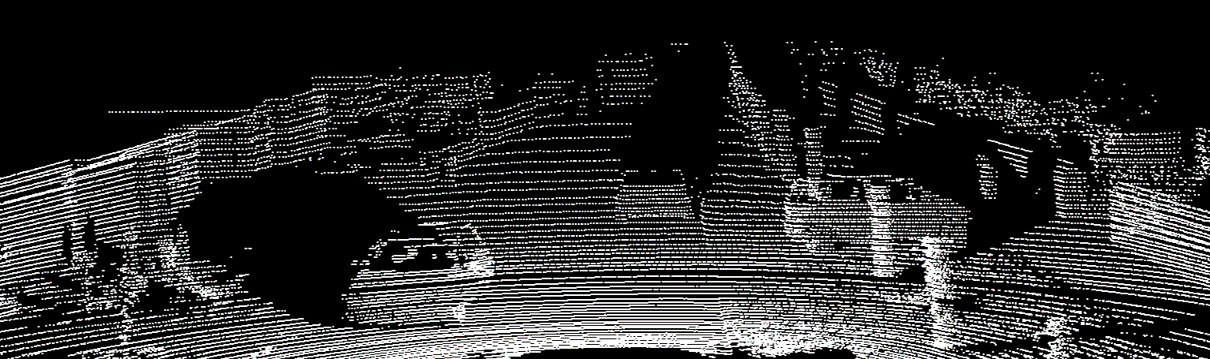}}
        \subfigure[]{\includegraphics[width=4.2cm]{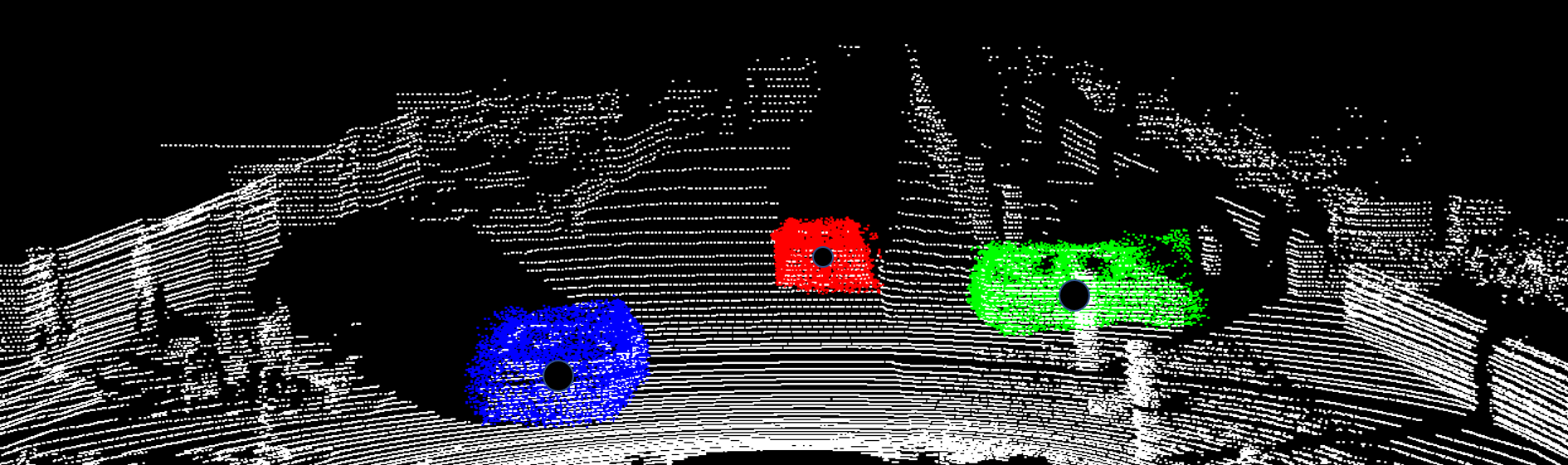}}
    \caption{(a) and (c) are RGB image and the corresponding 3D point cloud acquired by the camera and LiDAR. (b) and (d) are instance segmentation results of (a) and (c). Blue, red, green represents three different valid instances. Three black circles in (b) and (d) indicate instance centroids (IC) of each instance.}
    \label{IC}
\end{figure}

In SOIC, 2D Semantic Centroid (2D-SC) and 3D Semantic Centroid (3D-SC) are calculated using semantic segmentation of image and LiDAR point cloud. These two kinds of centroids are used to construct 2D-3D correspondences. Thus, the initialization problem of the LiDAR-Camera calibration is transformed into a PnP problem. Since the PnP algorithm needs at least three matching points, there must be three categories of objects in the current frame. SOIC chooses pedestrians, cars, and bicycles as the segmentation objects. However, in most cases, it is not easy to obtain all three categories of targets simultaneously. To solve this problem, we use instance segmentation to calculate the Instance Centroid (IC). Different from semantic segmentation, instance segmentation can distinguish different objects with the same category. Thus, we can obtain more 2D-3D correspondences than SOIC. An example of the 2D-IC and the 3D-IC is shown in Figure \ref{IC}.

We choose six categories as the valid instance: person, rider, car, truck, bus, and motorcycle. Given a frame of LiDAR point clouds ${{P}^{L}}$ that includes $N_{ins}^{L}$ valid instances, we define the 3D-IC of an instance $s\in \{1,2,\ldots ,N_{ins}^{L}\}$ as,
\begin{equation}
    IC_{s}^{L}=\frac{\sum\limits_{i=1}^{N_{s}^{L}}{P_{s,i}^{L}}}{N_{s}^{L}}
\end{equation}

\begin{equation}
    P_{s}^{L}=\{P_{s,1}^{L},P_{s,2}^{L},\ldots ,P_{s,N_{s}^{L}}^{L}|P_{s,i}^{L}\in {{P}^{L}},\ell _{i}^{P}=s\}
\end{equation}

where $P_{s}^{L}$ is the set of points with label $s$, $N_{s}^{L}$ is the total number of the points set, $\ell _{i}^{P}$ is the instance label that the 3D point $P_{s,i}^{L}$ belongs to. The definition of the 2D-IC is similar to the 3D-IC,
\begin{equation}
    IC_{s}^{I}=\frac{\sum\limits_{i=1}^{N_{s}^{I}}{p_{s,i}^{I}}}{N_{s}^{I}}
\end{equation}

\begin{equation}
    p_{s}^{I}=\{p_{s,1}^{I},p_{s,2}^{I},\ldots ,p_{s,N_{s}^{I}}^{I}|p_{s,i}^{I}\in I,\ell _{i}^{I}=s\}
\end{equation}

where $P_{s}^{I}$ is the set of pixels with label $s$, $N_{s}^{I}$ is the total number of the pixels set, $\ell _{i}^{I}$ is the instance label that the 2D pixel $p_{s,i}^{I}$ belongs to.

In most cases, the number of 2D instances (image) and 3D instances (point clouds) is different. The accurate matching relationship between the same instance is also unknown. We can match the 2D-IC and 3D-IC by sorting each instance with the help of the IC's location (point clouds: Y, image: u) and the number of instance's pixels or points. If the number of the 2D-3D correspondence $3\le {{N}_{2D-3D}}<5$, we adopt the P3P algorithm and EPnP algorithm within the RANSAC scheme when ${{N}_{2D-3D}}\ge 5$.

\section{Experiments and Discussion}\label{exp and dis}
\subsection{Dataset Preparation}\label{data pre}
We evaluate our approach on the KITTI benchmark dataset \cite{geiger2012we}, including RGB images and Velodyne point clouds recordings collected from different scenes. The timestamps of LiDAR and camera are synchronized, so the images and point clouds in each sequence correspond. To train our proposed calibration flow prediction network CFNet, we need to ensure the input, output, and corresponding calibration flow ground truth. We define the extrinsic parameters ground truth ${T}_{gt}$ between LiDAR and camera as the transformation matrix from the camera coordinate to the LiDAR coordinate. By adding a random variable $\Delta T$, we can obtain the initial calibration parameters ${{T}_{init}}=\Delta T\cdot {{T}_{gt}}$. The LiDAR point cloud is projected onto the image plane with initial extrinsic parameters ${T}_{init}$ and the camera intrinsic matrix $K$ to generate the LiDAR-image ${D}_{init}$. The network takes an RGB image $I$, and the corresponding projected LiDAR-image ${D}_{init}$ as input. The calibration flow ground truth can be provided by Eq. \ref{cf_gt}.

We use the odometry recordings from the KITTI dataset, specifically the left color image and Velodyne point clouds recordings. We use the sequence 06 to 21 for training and validation (29416 frames), sequence 01 to 05 for evaluation/test (4854 frames). The initial calibration off-range $\Delta T$ is $(\pm 1.5m,\pm 20{}^\circ )$. To compare with other learning-based (CNN-based) methods, we design four different test datasets on the raw recordings of the KITTI dataset. Each test dataset is independent of the training dataset with the following test name configurations:

\textbf{T1:} 0028 drive in $2009\_09\_30$ sequence, with initial off-range as $(\pm 1.5m,\pm 20{}^\circ )$.

\textbf{T2:} 0005/0070 drive in $2009\_09\_26$ sequence, with initial off-range as $(\pm 0.2m,\pm 15{}^\circ )$.

\textbf{T3:} 0005/0070 drive in $2009\_09\_26$ sequence, with initial off-range as $(\pm 0.2m,\pm 10{}^\circ )$.

\textbf{T4:} 0027 drive in $2009\_10\_03$ sequence, with initial off-range as $(\pm 0.3m,\pm 2{}^\circ )$.

Due to the image dimension in the KITTI benchmark is different (range from $\text{1224}\times \text{370}$ to $\text{1242}\times \text{376}$), pre-process is required to transform them to consistent size. To fulfill the input size of the network, that the input width and height is multiple of 32, we randomly crop the original image to  $\text{960}\times \text{320}$. We generate the sparse LiDAR-image by projecting the LiDAR point cloud onto the original image plane and then crop the original RGB image and the sparse LiDAR-image simultaneously. We can obtain the input data of the network without changing the camera intrinsic parameters. Besides, because the crop process is random, we can increase training data with this operation. Data augmentation is performed on the cropped input data. We add color augmentations with $50$ chance, where we perform random brightness, contrast, saturation, and hue shifts by sampling from uniform distributions in the ranges of $[0.7,1.3]$ for brightness, contrast, and saturation, $\left[ 1-0.3/3.14,1+0.3/3.14 \right]$ for hue shifts.

\subsection{Training Details}\label{training details}
For training the network, we use Adam Optimizer with an initial learning rate of $1{{e}^{-3}}$. We train our proposed calibration network on four Nvidia Titan RTX GPU with batch size 100 and total epochs 80. For the multi-range network, it is not necessary to retrain each network from scratch. Instead, a large-range model can be regarded as the pre-trained model for small-range training to speed up the training process. The training epoch of the model with the largest range is set to 80, while the others are set to 50. 

KITTI360 datasets are utilized as an additional dataset to test our proposed LiDAR-Camera calibration algorithm CFNet. The models trained on KITTI odometry training datasets are regarded as the pre-trained models. We only use sequences 0000, 0002, and 0003 for training and validation during training to fine-tune the CFNet models. Other sequences are selected as test datasets.

\subsection{Evaluation Metrics}\label{metrics}
We analyze the calibration results according to the rotation and the translation errors of the predicted extrinsic parameters. The translation error is evaluated by the Euclidian distance between the translation vectors. The absolute translation error is expressed as follows:

\begin{equation}
    {{E}_{t}}={{\left\| {{t}_{pred}}-{{t}_{gt}} \right\|}_{2}}
\end{equation}
where ${{\left\| \cdot  \right\|}_{2}}$ denotes the 2-norm of a vector. Besides, we also test the absolute error of the translation vector in $X,Y,Z$ directions respectively, ${T}_{X}, {T}_{Y}, {T}_{Z}$, and the mean value $\overline{t}=\left( {{E}_{X}}+{{E}_{Y}}+{{E}_{Z}} \right)/3$.

The rotation part is represented by quaternions that are transformed from the rotation matrix $R$. Since quaternion represents direction, we use quaternion angle distance to represent the difference between quaternions. The angle error of quaternion can be expressed as follows:
\begin{equation}
    {{E}_{R}}={{D}_{a}}({{q}_{gt}}*inv({{q}_{pred}}))
\end{equation}

\begin{equation}
    {{D}_{a}}(m)=atan2(\sqrt{b_{m}^{2}+c_{m}^{2}+d_{m}^{2}},\left| {{a}_{m}} \right|)
\end{equation}

\begin{equation}
    atan2(y,x)=\left\{\begin{array}{*{35}{l}}
    arctan(\tfrac{y}{x}) & x>0 \\
    arctan(\tfrac{y}{x})+\pi & x<0,y\ge 0 \\
    arctan(\tfrac{y}{x})-\pi & x<0,y<0 \\
    \frac{\pi }{2} & x=0,y>0 \\
    -\frac{\pi }{2} & x=0,y<0 \\
    undefined & x=0,y=0 \\
    \end{array} \right.
\end{equation}

where ${q}_{gt}$ is the ground truth quaternion , ${q}_{pred}$ is the predicted quaternion, $\left\{ {{a}_{m}},{{b}_{m}},{{c}_{m}},{{d}_{m}} \right\}$ is four components of the quaternion $m$, $*$ represents the quaternion multiplication, $inv$ represents the inverse of a quaternion. For a quaternion $m={{a}_{m}}+{{b}_{m}}i+{{c}_{m}}j+{{d}_{m}}k$, its inverse quaternion is expressed as:
\begin{equation}
    \begin{aligned}
      inv(m)=\frac{\overline{m}}{{{m}^{2}}} & =\frac{{{a}_{m}}-{{b}_{m}}i-{{c}_{m}}j-{{d}_{m}}k}{{{\left\| m \right\|}^{2}}} \\ 
      & =\frac{{{a}_{m}}-{{b}_{m}}i-{{c}_{m}}j-{{d}_{m}}k}{a_{m}^{2}+b_{m}^{2}+c_{m}^{2}+d_{m}^{2}} \\ 
    \end{aligned}
\end{equation}

where $\overline{m}$ is the conjugate of the quaternion $m$, $\left\| m \right\|$ is the length of quaternion. To test the angle error of the extrinsic rotation matrix on three degrees, we need to transform the rotation matrix to Euler angles and compute the angle error on Roll ${E}_{Roll}$, Pitch ${E}_{Pitch}$, Yaw ${E}_{Yaw}$, and the mean value $\overline{R}=\left( {{E}_{Roll}}+{{E}_{Pitch}}+{{E}_{Yaw}} \right)/3$.
\begin{equation}
    R_{pred}^{-1}\cdot {{R}_{gt}}=\left[ \begin{matrix}
    {{r}_{11}} & {{r}_{12}} & {{r}_{13}}  \\
    {{r}_{21}} & {{r}_{22}} & {{r}_{23}}  \\
    {{r}_{31}} & {{r}_{32}} & {{r}_{33}}  \\
    \end{matrix} \right]
\end{equation}

\begin{equation}
    \begin{aligned}
     & {{E}_{Yaw}}={{\theta }_{z}}=atan2({{r}_{21}},{{r}_{11}}) \\ 
     & {{E}_{Pitch}}={{\theta }_{y}}=atan2(-{{r}_{31}},\sqrt{r_{31}^{2}+r_{33}^{2}}) \\ 
     & {{E}_{Roll}}={{\theta }_{x}}=atan2({{r}_{32}},{{r}_{33}}) \\ 
    \end{aligned}
\end{equation}

Besides, we apply another two $se(3)$ error based evaluation metrics mentioned in \cite{yuan2020rggnet}: Mean $se(3)$ error (MSEE) and Mean re-calibration rate (MRR).

\subsection{Results and discussion}\label{results}
\begin{figure}[thpb]
    \centering
    \includegraphics[width=8.5cm]{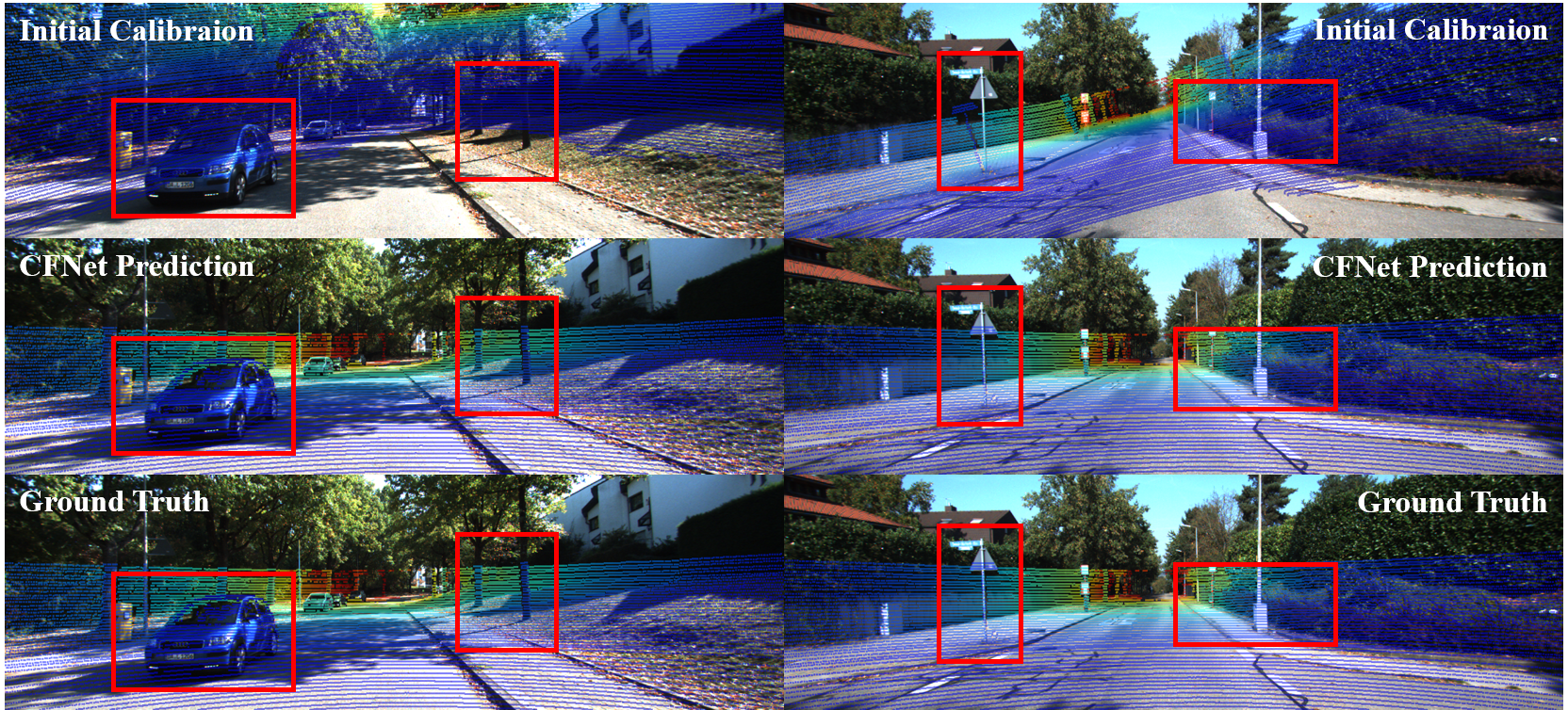}
    \caption{Examples of CFNet predictions on the KITTI odometry test datsets. Reference objects for calibration are shown in the red bounding boxes}
    \label{Figure_odom}
\end{figure}

The evaluation results on the KITTI odometry dataset are shown in Table \ref{Results_odom}. It can be seen that in all of these test sequences, the mean translation error ${E}_{t}<2cm$ and the mean rotation error ${E}_{R}<0.13{}^\circ$. Figure \ref{Figure_odom} shows two examples of CFNet predictions. We can see that the reference objects in the projected depth image and RGB image align accurately after re-calibration. In all of these test sequences, the calibration error of sequence 01 is the largest. The main reason is that this sequence is collected from a high-way scene, which is not included in the training dataset.

Table \ref{Results_odom_init} and Table \ref{Comparison_odom_init} illustrate the calibration results with the semantic initialization. Compared to random initial calibration parameters, the semantic method can provide better initial parameters. Therefore, the final calibration error is much smaller. Furthermore, the semantic initialization procedure is automatic, and the basic requirement is the number of the valid instance correspondences in the scene ${{N}_{2D-3D}}\ge 3$. As shown in Table \ref{Comparison_odom_init}, the translation error of semantic initialization is much bigger than the rotation error. Our proposed CFNet can also predict accurate calibration parameters, even if the initial parameters provided by semantic initialization have a big deviation from the ground truth. 

\begin{figure}[thpb]
    \centering
    \includegraphics[width=8.5cm]{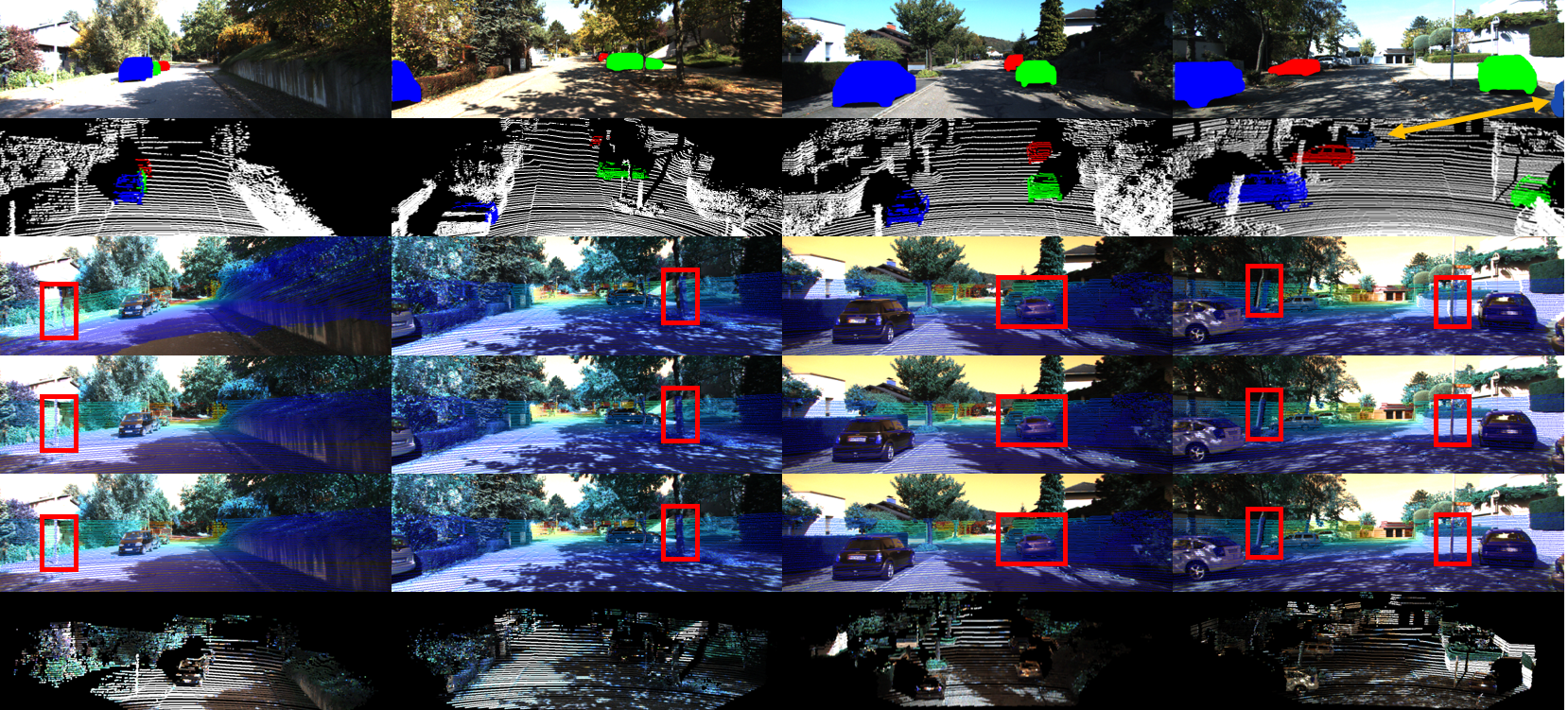}
    \caption{Examples of CFNet predictions with semantic initialization.(First Row) 2D-IC. (Second Row) 3D-IC. (Third Row) Semantic Initialization. (Fourth Row) CFNet Prediction. (Fifth Row) Ground Truth. (Bottom Row) Rectified calibrated point cloud with RGB color.}
    \label{Figure_odom_init}
\end{figure}

Figure \ref{Figure_odom_init} shows some examples of CFNet predictions with semantic initialization. There are many occlusions between the detected instances in the first column, so the 2D-3D instance matching is not accurate, and the predicted initial calibration parameters have a large deviation. Instance occlusion in the second column image is better than that in the first column image, so the initial error is much smaller. The 2D and 3D instances detected in the third column are complete without any occlusion, so the initial calibration parameters are very accurate, and the deviation from the ground truth is tiny. In the last column, we find a mismatch between the 2D and 3D instances (represented by an arrow), but we can still get an excellent initial calibration value. CFNet can re-calibrate accurately no matter the initial calibration error is large or small by using semantic initialization.

\begin{table*}[]
\caption{The results of the KITTI odometry test sequences with initial off-range as $(\pm 1.5m,\pm 20{}^\circ )$}
\label{Results_odom}
\begin{center}
\begin{tabular}{cccccccccc}
\toprule
\multicolumn{2}{c}{\multirow{2}{*}{Sequence}} & \multicolumn{4}{c}{Translation   (cm)}                   & \multicolumn{4}{c}{Rotation (${}^\circ   $)}                                                 \\
\multicolumn{2}{c}{}                          & ${{E}_{t}}$    & ${{E}_{X}}$ & ${{E}_{Y}}$ & ${{E}_{Z}}$ & ${{E}_{R}}$                  & ${{E}_{Roll}}$                  & ${{E}_{Pitch}}$           & ${{E}_{Yaw}}$ \\
\midrule
\multirow{3}{*}{01}          & Mean           & 1.994 & 1.006       & 1.220       & 1.538       & 0.129 & 0.049                   & 0.078                    & 0.047                  \\
                             & Median         & 1.830          & 1.046       & 1.068       & 1.619       & 0.120          & 0.041                   & 0.069                    & 0.032                  \\
                             & Std.           & 1.027          & 0.440       & 0.462       & 0.850       & 0.065          & 0.028                   & 0.080                    & 0.028                  \\
\multirow{3}{*}{02}          & Mean           & 1.773          & 0.134       & 1.334       & 0.365       & 0.094          & 0.073                   & 0.085                    & 0.054                  \\
                             & Median         & 1.573          & 0.115       & 1.561       & 0.367       & 0.086          & 0.093                   & 0.085                    & 0.067                  \\
                             & Std.           & 1.051          & 0.067       & 0.421       & 0.153       & 0.050          & 0.043                   & 0.074                    & 0.030                  \\
\multirow{3}{*}{03}          & Mean           & 1.935          & 0.527       & 0.948       & 0.776       & 0.115          & 0.078                   & 0.125                    & 0.074                  \\
                             & Median         & 1.807          & 0.512       & 0.830       & 0.696       & 0.106          & 0.081                   & 0.106                    & 0.074                  \\
                             & Std.           & 1.411          & 0.496       & 0.212       & 0.184       & 0.057          & 0.060                   & 0.058                    & 0.044                  \\
\multirow{3}{*}{04}          & Mean           & 1.748          & 0.927       & 0.675       & 0.552       & 0.075          & 0.031                   & 0.030                    & 0.019                  \\
                             & Median         & 1.585          & 1.128       & 0.494       & 0.381       & 0.073          & 0.027                   & 0.026                    & 0.016                  \\
                             & Std.           & 0.958          & 0.787       & 0.549       & 0.460       & 0.030          & 0.015                   & 0.022                    & 0.012                  \\
\multirow{3}{*}{05}          & Mean           & 1.766          & 0.723       & 0.410       & 0.848       & 0.121          & 0.010                   & 0.017                    & 0.015                  \\
                             & Median         & 1.606          & 0.038       & 0.490       & 0.489       & 0.112          & 0.010                   & 0.021                    & 0.015                  \\
                             & Std.           & 0.925          & 1.219       & 0.363       & 0.924       & 0.062          & 0.007                   & 0.011                    & 0.009         \\ 
\bottomrule
\end{tabular}
\end{center}
\end{table*}


\begin{table*}[]
\caption{The results of the KITTI odometry test sequences with semantic initialization}
\label{Results_odom_init}
\begin{center}
\begin{tabular}{cccccccccc}
\toprule
\multicolumn{2}{c}{\multirow{2}{*}{Sequence}} & \multicolumn{4}{c}{Translation   (cm)}                & \multicolumn{4}{c}{Rotation (${}^\circ   $)}                   \\
\multicolumn{2}{c}{}                          & ${{E}_{t}}$ & ${{E}_{X}}$ & ${{E}_{Y}}$ & ${{E}_{Z}}$ & ${{E}_{R}}$ & ${{E}_{Roll}}$ & ${{E}_{Pitch}}$ & ${{E}_{Yaw}}$ \\
\midrule
\multirow{3}{*}{02}          & Mean           & 1.724       & 0.687       & 0.217       & 0.538       & 0.171       & 0.064          & 0.030           & 0.084         \\
                             & Median         & 1.692       & 0.211       & 0.179       & 0.522       & 0.147       & 0.090          & 0.023           & 0.076         \\
                             & Std.           & 0.758       & 0.995       & 0.144       & 0.451       & 0.101       & 0.050          & 0.016           & 0.015         \\
\multirow{3}{*}{03}          & Mean           & 1.835       & 1.459       & 0.654       & 0.638       & 0.133       & 0.056          & 0.044           & 0.043         \\
                             & Median         & 1.930       & 1.575       & 0.648       & 0.804       & 0.110       & 0.044          & 0.034           & 0.018         \\
                             & Std.           & 0.668       & 0.853       & 0.853       & 0.360       & 0.081       & 0.037          & 0.038           & 0.060         \\
\multirow{3}{*}{05}          & Mean           & 1.711       & 0.928       & 0.523       & 0.657       & 0.123       & 0.084          & 0.090           & 0.044         \\
                             & Median         & 1.772       & 1.243       & 0.165       & 0.423       & 0.113       & 0.071          & 0.050           & 0.034         \\
                             & Std.           & 0.748       & 0.575       & 0.657       & 0.701       & 0.054       & 0.070          & 0.088           & 0.045      \\  
\bottomrule
\end{tabular}
\end{center}
\end{table*}

\begin{table*}[]
\caption{The Comparison results of semantic initialization and CFNet prediction}
\label{Comparison_odom_init}
\begin{center}
\begin{tabular}{ccccccccc}
\toprule
\multirow{3}{*}{Sequence} & \multicolumn{4}{c}{Semantic   Initialization}                                           & \multicolumn{4}{c}{CFNet}                                                               \\
                          & \multicolumn{2}{c}{Translation   (cm)} & \multicolumn{2}{c}{Rotation   (${}^\circ   $)} & \multicolumn{2}{c}{Translation   (cm)} & \multicolumn{2}{c}{Rotation   (${}^\circ   $)} \\
\midrule
                          & ${{E}_{t}}$      & $\overline{t}$\tnote{a}      & ${{E}_{R}}$          & $\overline{R}$\tnote{b}          & ${{E}_{t}}$      & $\overline{t}$\tnote{a}      & ${{E}_{R}}$          & $\overline{R}$\tnote{b}          \\
02                        & 60.405           & 50.568              & 1.753                & 0.960                   & 1.724            & 0.481               & 0.171                & 0.059                   \\
03                        & 19.148           & 11.257              & 0.301                & 0.159                   & 1.835            & 0.917               & 0.133                & 0.048                   \\
05                        & 9.092            & 7.438               & 1.728                & 0.168                   & 1.711            & 0.702               & 0.123                & 0.073           \\       
\bottomrule
\end{tabular}
\end{center}
\end{table*}

The calibration results on the KITTI raw recordings are shown in Table \ref{raw_T1} and Table \ref{raw_T234}. As shown in Table \ref{raw_T1}, the CFNet achieves a mean translation error of 0.995cm (X, Y, Z: 1.025cm, 0.919cm, 1.042cm) and a mean angle error of $0.087{}^\circ$(Roll, Pitch, Yaw: $0.059{}^\circ ,\text{ }0.110{}^\circ ,\text{ }0.092{}^\circ $). It is obvious that the CFNet is far superior to RegNet and CalibNet. Although the mis-calibration range of the CalibNet $(\pm 0.2m,\pm 20{}^\circ )$ is smaller than CFNet $(\pm 1.5m,\pm 20{}^\circ )$, the calibration performance of CFNet is much better than CalibNet.

The comparison results on the test datasets T2, T3, and T4 shown in Table \ref{raw_T234} also illustrate the superior of the CFNet. Compared to the translation errors and the rotation errors utilized in the above experiments, $se3$ error \cite{yuan2020rggnet} is a more direct evaluation metric. The off-range of the test dataset T3 is a litter smaller than that in dataset T2. The metric MSEE for $\beta$-RegNet and RGGNet is smaller in T3 than in T2. Nevertheless, the MSEE is the same for CFNet in these two test datasets, which proves that CFNet is more robust than $\beta$-RegNet and RGGNet with different off-range settings. In test dataset T4, the performance of $\beta$-RegNet degrades heavily, and RGGNet needs to re-train on an additional dataset by adding a small number of data from $2009\_10\_03$ sequence to achieve a good calibration result $0.010\left( 83.22\% \right)$. CFNet does not need any additional training dataset and re-train process. The calibration results $0.001\left( 98.08\% \right)$ demonstrates that CFNet has good generalization capability. Thus, FNet outperforms all of the learning-based calibration algorithms, RegNet, CalibNet, RGGNet, and even the motion-based calibration method. We can also see that, compared to the motion-based algorithm \cite{taylor2016motion}, our proposed method is more generalized, without the requirements of hand-crafted features or the extra IMU sensor. Furthermore, the semantic initialization process does not require motion information which is vital to the hand-eye calibration. 

\begin{table*}[]
\caption{The comparison results on the T1 test dataset}
\label{raw_T1}
\begin{center}
\begin{tabular}{cccccccccc}
\toprule
\multirow{3}{*}{Methods} & \multirow{3}{*}{Mis-calibration Range} & \multicolumn{8}{c}{T1}                                                                                                       \\
                         &                                        & \multicolumn{4}{c}{Translation (cm)}                     & \multicolumn{4}{c}{Rotation (${}^\circ $)}                        \\
                         &                                        & ${{E}_{X}}$ & ${{E}_{Y}}$ & ${{E}_{Z}}$ & $\overline{t}$\tnote{a} & ${{E}_{Roll}}$ & ${{E}_{Pitch}}$ & ${{E}_{Yaw}}$ & $\overline{R}$\tnote{b} \\
\midrule
RegNet \cite{schneider2017regnet}                   & $(\pm 1.5m,\pm 20{}^\circ )$           & 7           & 7           & 4           & 6              & 0.24           & 0.25            & 0.36          & 0.28           \\
CalibNet \cite{iyer2018calibnet}                 & $(\pm 0.2m,\pm 20{}^\circ )$           & 4.2         & 1.6         & 7.22        & 4.34           & 0.15           & 0.9             & 0.18          & 0.41           \\
CFNet                    & $(\pm 1.5m,\pm 20{}^\circ )$           & \textbf{1.025}       & \textbf{0.919}       & \textbf{1.042}       & \textbf{0.995}          & \textbf{0.059}          & \textbf{0.110}           & \textbf{0.092}         & \textbf{0.087}  \\
\bottomrule
\end{tabular}
\end{center}
\end{table*}

\begin{table*}[]
\caption{The comparison results on the T2, T3, and T4 test dataset}
\label{raw_T234}
\begin{center}
\begin{threeparttable}
\begin{tabular}{cccc}
\toprule
MSEE(MRR)      & T2             & T3             & T4             \\
\midrule
TAYLOR \cite{taylor2016motion}         & *              & *              & 0.010($\dagger$)       \\
CalibNet \cite{iyer2018calibnet}       & -              & 0.022(-)       & *              \\
$\beta$-RegNet \cite{yuan2020rggnet} & 0.048(53.23\%) & 0.046(34.14\%) & 0.092(-1.89\%) \\
RGGNet \cite{yuan2020rggnet}         & 0.021(78.40\%) & 0.017(72.64\%) & 0.010(83.22\%) \\
CFNet          & \textbf{0.003(96.74\%)} & \textbf{0.003(94.53\%)} & \textbf{0.001(98.08\%)} \\
\bottomrule
\end{tabular}
\begin{tablenotes}
\footnotesize
\item[] * means the author does not provide the result in the paper.
\item[] - shows that the calibration algorithm fails.
\item[] $\dagger$ represents that the calibration algorithm does not have this metrics.
\end{tablenotes}
\end{threeparttable}
\end{center}
\end{table*}

\begin{figure}[thpb]
    \centering
    \includegraphics[width=8.5cm]{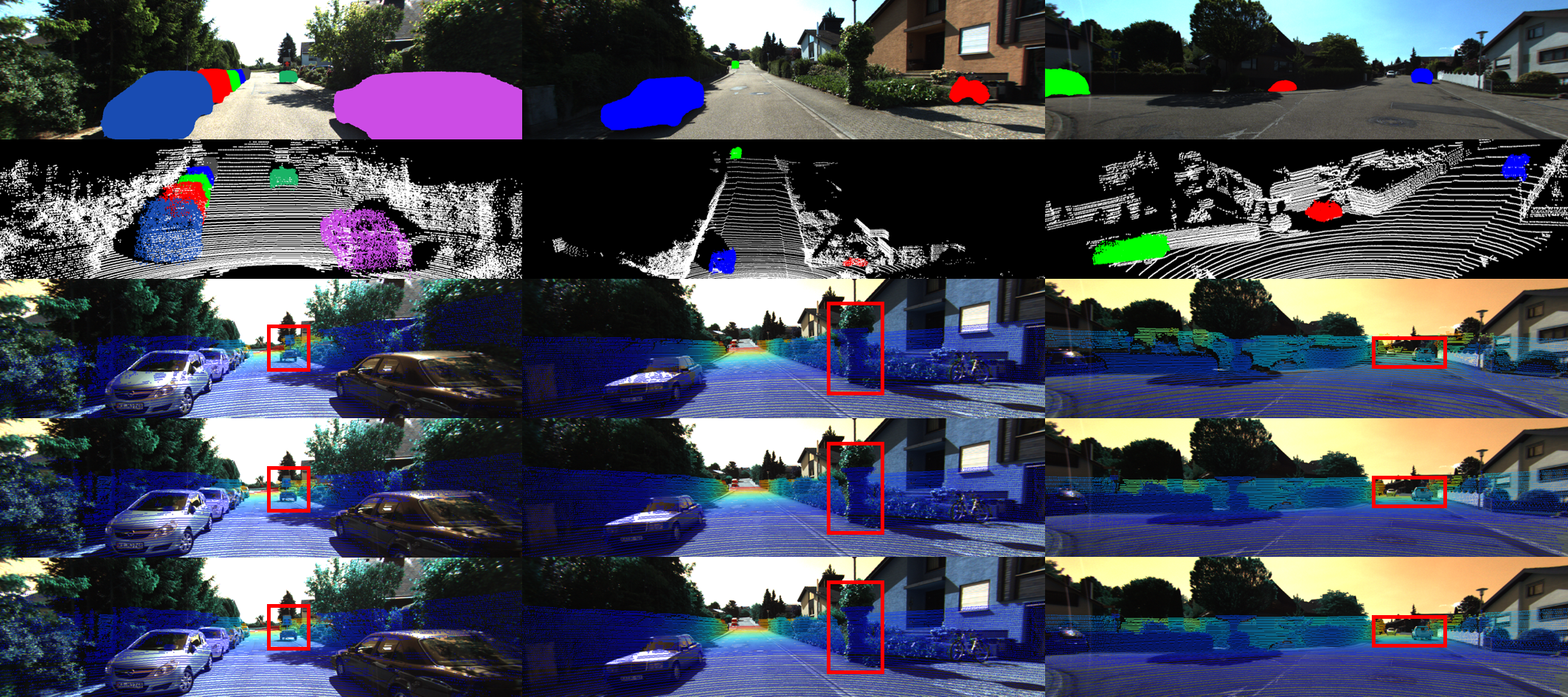}
    \caption{Examples of CFNet predictions with semantic initialization on the KITTI360 datasets.(First Row) 2D-IC. (Second Row) 3D-IC. (Third Row) Semantic Initialization. (Forth Row) CFNet Prediction. (Fifth Row) Ground Truth.}
    \label{Figure_360_init}
\end{figure}

We also test the performance of CFNet on the KITTI360 benchmark dataset. The results are shown in Table \ref{KITTI360} and Figure \ref{Figure_360_init}. Despite re-training on a tiny sub dataset with one epoch, excellent results are obtained in the test sequences. Therefore, when the sensor parameters change, such as the camera focal length or the LiDAR-Camera extrinsic parameters, an excellent prediction model can be obtained with simple re-training.

\begin{table*}[]
\caption{The calibration results on KITTI360 test dataset}
\label{KITTI360}
\begin{center}
\begin{tabular}{ccccc}
\toprule
\multirow{3}{*}{Sequence} & \multicolumn{4}{c}{CFNet}                                                             \\
                          & \multicolumn{2}{c}{Translation   (cm)} & \multicolumn{2}{c}{Rotation (${}^\circ   $)} \\
                          & ${{E}_{t}}$      & $\overline{t}$      & ${{E}_{R}}$         & $\overline{R}$         \\
\midrule
0003                      & 1.941            & 0.656               & 0.137               & 0.031                  \\
0004                      & 2.296            & 0.661               & 0.335               & 0.159                  \\
0005                      & 2.099            & 0.594               & 0.260               & 0.095                  \\
0006                      & 2.341            & 0.790               & 0.224               & 0.081                  \\
0007                      & 2.881            & 1.207               & 0.280               & 0.119                  \\
0009                      & 2.179            & 1.025               & 0.171               & 0.075                  \\
0010                      & 2.485            & 0.999               & 0.211               & 0.073             \\    
\bottomrule
\end{tabular}
\end{center}
\end{table*}

\begin{figure*}[thpb]
    \centering
        \subfigure[$2013\_05\_28\_drive\_0005\_sync$]{\includegraphics[width=8.5cm]{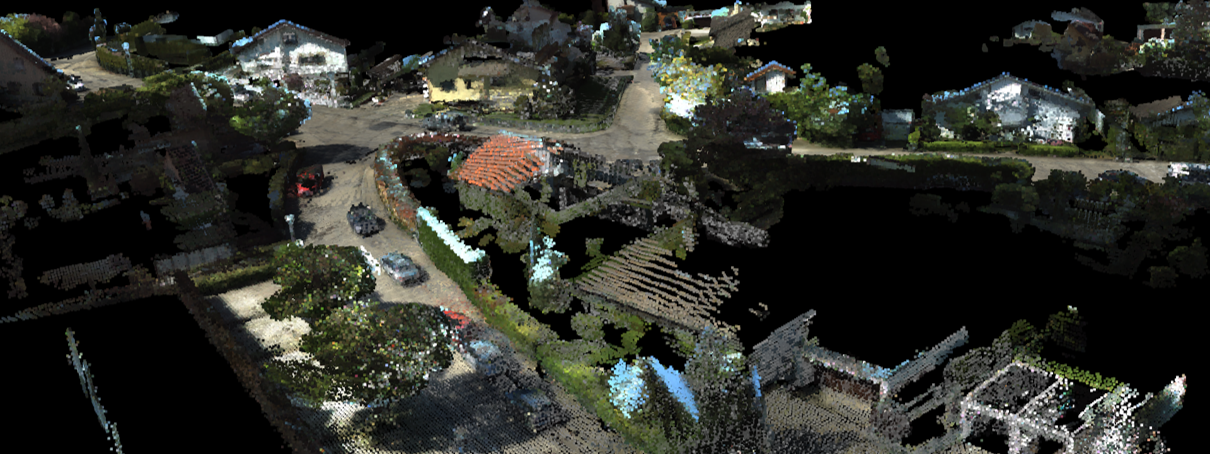}} 
        \subfigure[$2013\_05\_28\_drive\_0010\_sync$]{\includegraphics[width=8.5cm]{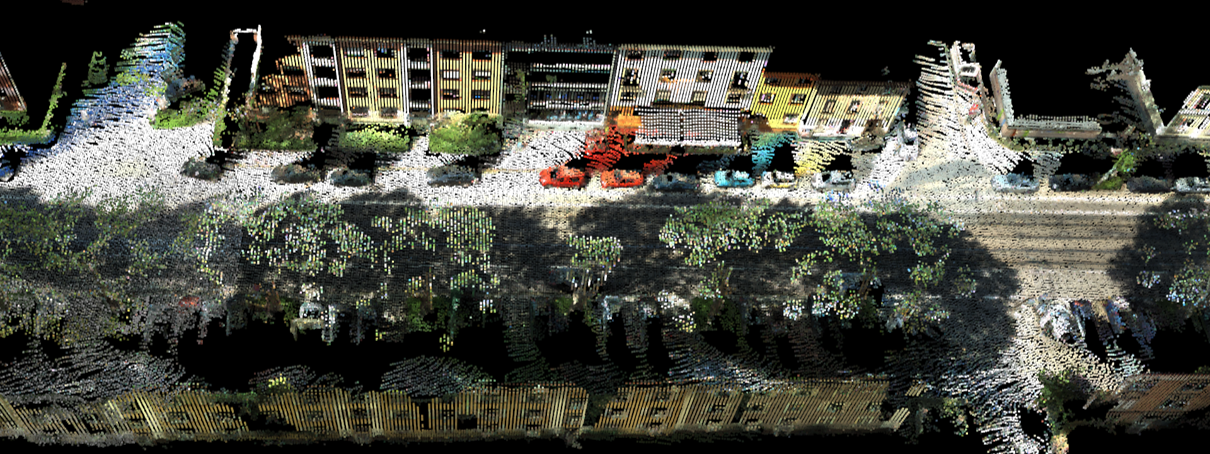}}
    \caption{Examples of the reconstructed 3D color map on the KITTI360 datasets with the prediciton of CFNet.}
    \label{Map_360}
\end{figure*}

\section{Conclusion}
In this paper, we presented a novel online LiDAR-camera extrinsic calibration algorithm. To represent the deviation from the initial projection of LiDAR point clouds to the ground truth, we define an image called calibration flow. Inspired by the optical flow network, we design a deep calibration flow network CFNet. The initial projected points are rectified by the prediction of CFNet to construct accurate 2D-3D correspondences. EPnP algorithm within the RANSAC scheme is utilized to estimate the extrinsic parameters with iterative refinement. Our experiments demonstrate the superiority of CFNet. The additional experiments on the KITTI360 datasets illustrate the generalization of our method. 

Moreover, we propose a semantic initialization algorithm based on the instance centroids. Unlike existing initialization techniques, we do not rely on motion information or manual measurement. This initialization procedure is effective when the scene contains enough valid instances.


%







\bibliographystyle{IEEEtran}
\bibliography{IEEEabrv, reference}
\end{document}